\documentclass[10pt,twocolumn,letterpaper]{article}
\usepackage{arxiv}
\usepackage{times}
\usepackage{epsfig}
\usepackage{graphicx}
\usepackage{amsmath}
\usepackage{amssymb}

\usepackage{amsfonts}
\usepackage{bm}
\usepackage[dvipsnames]{xcolor}
\usepackage{multirow}
\usepackage{booktabs}

\usepackage{setspace}
\usepackage{verbatim}
\usepackage{tabularx}

\usepackage{soul}
\usepackage{url}
\usepackage[utf8]{inputenc}
\usepackage{amsthm}
\usepackage{algorithm}
\usepackage{adjustbox}
\usepackage[symbol]{footmisc}

\usepackage[numbers]{natbib}

\newcommand{\nibf}[1]{\noindent\textbf{#1}\ }

\newcommand{\prop}{LatteGAN}
\newcommand{\propL}{Visually Guided Language Attention GAN}

\newcommand{\txtcond}{Text-Conditioned U-Net discriminator}
\newcommand{\task}{GeNeVA}
\newcommand{\geneva}{GeNeVA}    

\newcommand{\rone}[2]{#2}
\newcommand{\rtwo}[2]{#2}

\newcommand{\sepmode}{}

\usepackage[pagebackref=true,breaklinks=true,colorlinks,bookmarks=false]{hyperref}

\arxivfinalcopy 

\ifarxivfinal\fi

\begin{document}

\title{
\prop{}: Visually Guided Language Attention
for Multi-Turn Text-Conditioned Image Manipulation
}

\author{Shoya Matsumori
\thanks{Corresponding author}
\thanks{Equally contributed}
,
Yuki Abe\textsuperscript{\dag{}},
Kosuke Shingyouchi,\\
Komei Sugiura,
Michita Imai\\ 
Keio University\\
{\tt\small shoya@ailab.ics.keio.ac.jp}
}

\maketitle

\begin{abstract}
Text-guided image manipulation tasks have recently gained attention in the vision-and-language community. While most of the prior studies focused on single-turn manipulation, our goal in this paper is to address the more challenging multi-turn image manipulation (MTIM) task. Previous models for this task successfully generate images iteratively, given a sequence of instructions and a previously generated image. However, this approach suffers from under-generation and a lack of generated quality of the objects that are described in the instructions, which consequently degrades the overall performance. To overcome these problems, we present a novel architecture called a \propL{}~(\prop{}). Here, we address the limitations of the previous approaches by introducing a Visually Guided Language Attention (Latte) module, which extracts fine-grained text representations for the generator, and a \txtcond{} architecture, which discriminates both the global and local representations of fake or real images. Extensive experiments on two distinct MTIM datasets, CoDraw and i-CLEVR, demonstrate the state-of-the-art performance of the proposed model. 
\rtwo{1}{The code is available online (\url{https://github.com/smatsumori/LatteGAN}).}
\end{abstract}

\begin{figure}[t]
\begin{center}
  \includegraphics[clip,width=\linewidth]{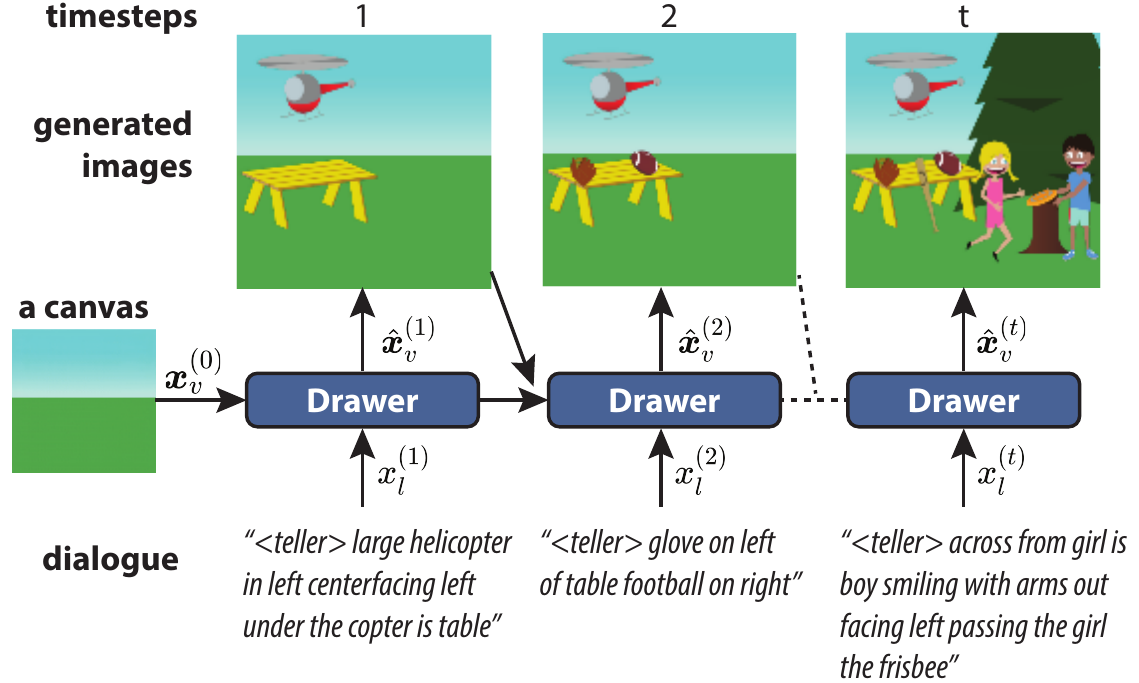}
  \caption{
  \textbf{Overview of the Generative Neural Visual Artist (\geneva{}) task.}
  \rone{1}{In this task, a Drawer iteratively modifies an image according to a sequence of instructions given by a Teller. The objective of this task is to build a Drawer model that can perform pixel-wise generation of an image. }
  }\label{fig:task}
  \end{center}
\end{figure}
\section{Introduction}
Recent advancements in deep generative models~\citep{kingma2013auto,goodfellow2014generative} have pushed the boundary of limitations in \rone{7}{vision and language} studies to 
\rtwo{4}{a great extent~\citep{ma2019unpaired}}.
Text-guided image generation tasks~\citep{reed2016generative,xu2018attngan} are among the prominent topics of study in this discipline. This topic has a wide range of applications spanning education, entertainment, computer-supported design, and human-robot collaboration. In particular, in a human-robot interaction scenario~\citep{kanda2004development}, this technology enables the visualization of the consequence of future actions and the internal beliefs of a robot~\citep{magassouba2018multimodal}.

One of the most prominent studies on interactive text-based image manipulation was by~\citet{el2019tell}.
This research introduced a multi-turn text-conditioned image manipulation (MTIM) task, namely,
\rone{1}{the GeNeVA task (Fig.~\ref{fig:task}), whose objective is to iteratively manipulate an image according to a given sequence of instructions and a previously generated image.}
The GeNeVA-GAN architecture they proposed  involves recurrent structures in which objects are successfully generated on a background and then simple transformations are applied to existing objects. However, that approach suffers from the under-generation of instructed objects, resulting in a low recall rate. This is especially problematic for iterative manipulation, since instructions often refer to previously generated images.

\begin{figure}[tbp]
\centering
  \includegraphics[clip,width=\linewidth]{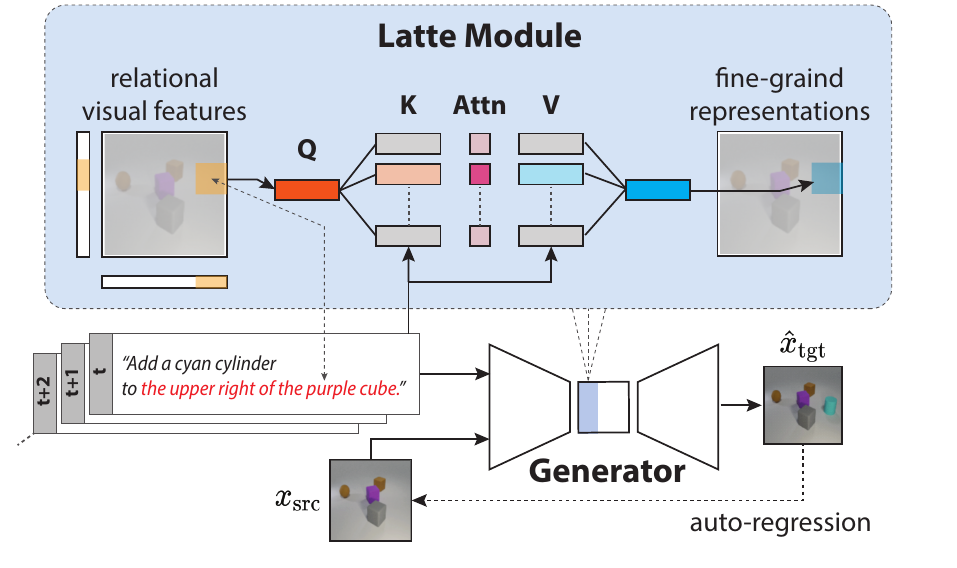}
  \caption{\textbf{Overview of the Latte module}. LatteGAN iteratively generates images according to a sequence of instructions. The Visually Guided Language Attention (Latte) module uses the relational features as queries to extract fine-grained representations from each instruction. }
  	\label{fig:1}
\end{figure}

To overcome that problem, we propose a \propL{}~(\prop{}), a novel Generative Adversarial Network~(GAN) architecture for multi-turn text-conditioned image manipulation, which is accompanied by a Visually Guided Language Attention (Latte) module, as shown in Fig.~\ref{fig:1}. The Latte module exploits a source-target attention structure that takes word embeddings as keys and values and takes relational visual representations as queries. With this structure, spatially variant fine-grained instruction representations, which store a specific instruction feature at a specific location, can be extracted. Additionally, a Text-Conditioned U-Net discriminator is introduced. This approach enables both global- and local-level discrimination of images by using a global text-conditioned loss to verify whether an image was modified as instructed and a local unconditioned loss to evaluate the image at the local level.

\rtwo{2}{In summary, the contributions of this paper are twofold:}
\begin{itemize}
 \item We propose \propL{}~(\prop{}), a multi-turn text-conditioned image generation GAN accompanied by two key components: (1) a Latte module that can extract the fine-grained instruction representations that are crucial for image modification; and (2) a Text-Conditioned U-Net discriminator that can discriminate images on the basis of  both their modification and their quality.
 \item \rtwo{2}{We conducted experiments on two different MTIM tasks: CoDraw~\citep{kim2019codraw} and i-CLEVR~\citep{el2019tell}. Our experimental results demonstrate that the proposed approach can outperform the baseline method, showing significant effectiveness in the multi-turn text-conditioned image generation task. Additionally, the effectiveness of our approach was validated by ablation studies.}
\end{itemize}

\sepmode{}
\section{Related Works}
\nibf{Conditional image generation.} There is an extensive amount of research in the field of image synthesis using GANs~\citep{goodfellow2014generative}. Conditional image generation, an approach that takes extra inputs to guide the output, is one of the most studied topics in this field. Various kinds of conditioning have been examined, such as discrete labels~\citep{mirza2014conditional}, images~\citep{isola2017image}, and masks~\citep{wang2018high}.

\nibf{Text-conditioned image generation.} \rtwo{4}{This setting, which takes a natural language text as an input, has also gained attention recently~\cite{tan2019semantics,li2020exploring,yuan2019ckd}.}
One of the first attempts to use GAN models in this field was presented in the work by~\citet{reed2016generative}. Following that, \citet{zhang2017stackgan} proposed StackGAN, which uses two-stage generation consisting of low-resolution generation and high-resolution refinement. \citet{hong2018inferring} presented a three-stage generation approach consisting of bounding-box generation, mask generation, and image generation. \rone{8}{\citet{xu2018attngan} proposed AttnGAN}, which incorporates an attention module using fine-grained representations obtained from word embeddings in addition to  a text embedding.

\nibf{Text-conditioned image manipulation.} This approach aims to semantically  manipulate an image rather than create it from scratch. Similarly to \citet{xu2018attngan}, \citet{nam2018text} introduced TAGAN, which includes word-level local discriminators. ManiGAN~\cite{li2020manigan} used two modules to manipulate only related regions that match text descriptions. Another branch of research aims to perform manipulations based on complex instructions~\citep{kenan2020learning,zhang2020text} that include specifications for the target location (\textit{where}), the target object (\textit{what}), and the target operation type (\textit{how}).

\nibf{Multi-turn text-conditioned image manipulation.} While most of the above studies are related to text-conditioned image manipulation concentrated on a single turn of generation, our research target is multi-turn text-guided image generation (MTIM). \citet{el2019tell} first introduced the MTIM task known as Generative Neural Visual Artist (\geneva{}; Fig.~\ref{fig:task}), accompanied by two distinct datasets, CoDraw~\citep{kim2019codraw} and i-CLEVR~\citep{el2019tell}. \citet{fu2020iterative} introduced the self-supervised counterfactual reasoning~(SSCR) framework to overcome data scarcity in the MTIM task.

\nibf{Limitations.} Although the previous works have successfully pioneered this field of study, some limitations remain to be addressed. The most prominent is the fact that current methods  often overlook manipulation instructions and fail to generate objects. This is particularly problematic for the MTIM task, as generating an image at a certain step often involves reference to the previous images.

\sepmode{}
\section{Task Specifications}
\begin{figure*}[th]
  \includegraphics[clip,width=\linewidth]{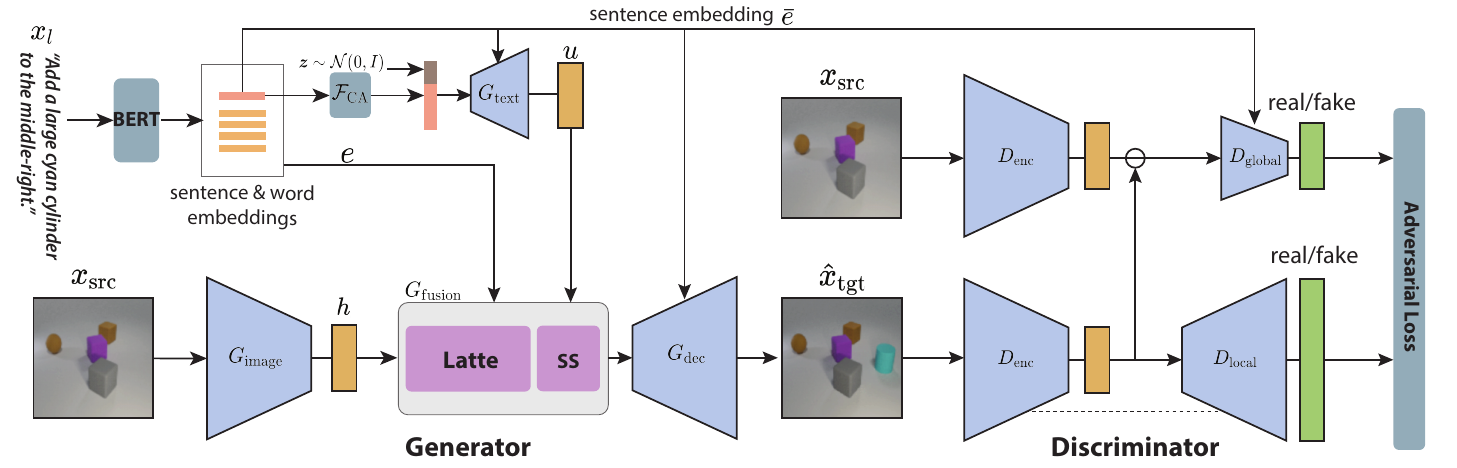}
  \caption{\textbf{Architecture of \prop{}}. For each timestep $t$, the generator takes as inputs a text instruction $\bm{x}_{l}$ and a previous generated image $\bm{x}_{\mathrm{src}}$, and generates a target image $\hat{\bm{x}}_{\mathrm{tgt}}$. For readability, notations for timestep $t$ are omitted in the figure. $\mathcal{F}_{\mathrm{CA}}$ denotes the conditional augmentor and SS denotes the semantic synthesis module. The discriminator judges whether the generated image is real or fake. Here, two types of discriminators are introduced: a local level unconditional discriminator and a global level text-conditioned discriminator. } 
  	\label{fig:model}
\end{figure*}

The \geneva{} task is a multi-turn text-conditioned image generation (MTIM) task. It involves two participants: a Teller that instructs how to modify the image, and a Drawer that draws the image according to the Teller's instructions. The final goal of this task is to train the Drawer model. Fig. ~\ref{fig:task} shows typical scenes in the \task{} task. Given an image input $\hat{\bm{x}}_{v}^{(t-1)}$ and an instruction $x_{l}^{(t)}$, the Drawer generates an image $\hat{\bm{x}}_{v}^{(t)}$. 
\rone{5}{In the first turn of generation, an empty canvas with a background image $\bm{x}_{v}^{(0)}$ is given to the Drawer. In the following turns, the Drawer takes as input its own previously generated image $\hat{\bm{x}}_{v}^{(t-1)}$.}

To achieve this goal, the Drawer model needs to fulfill the following three requirements. (1) It must comprehend the natural language instructions and map them to the actual canvas. This process includes visually grounded language understanding, such as referring expression comprehension (e.g., ``Put a pine tree to the left of the girl''). Additionally, the model has to (2) perform pixel-based image generation according to instructions. This may involve training a generative model such as a GAN. Finally, the model needs to (3) generate images several times in accordance with a dialogue. Unlike in previous text-conditioned image generation tasks, the model is supposed to generate several  intermediate results in an auto-regressive manner.

\sepmode{}
\section{\prop{}}
\nibf{Architecture.} Fig.~\ref{fig:model} illustrates the overview of \prop{}. Its main components are (1) the Latte module in the generator and (2) the Text Conditioned U-Net discriminator. (1) The Latte module consists of a Relational Visual Feature Extractor, which extracts relational visual features from the source image, and a source-target-attention structure, which attends  to the necessary instruction text tokens queried via the relational visual features. This structure efficiently extracts fine-grained features. (2) The Text-Conditioned U-Net discriminator consists of a global text-conditioned discriminator and a local-level unconditioned discriminator. By globally discriminating the manipulation based on the instructions and locally discriminating the image, this structure verifies that the generated image is based on the instructions and has high quality.

\nibf{Inputs and output.} At timestep $t$, there are two inputs to the model: an image input $\bm{x}_{\mathrm{src}} = \hat{\bm{x}}_{v}^{(t-1)}$ and a raw-text input $x_{l} = x_{l}^{(t)}$. The output is a generated image $\hat{\bm{x}}_{\mathrm{tgt}} = \hat{\bm{x}}_{v}^{(t)}$. The transition between timesteps $t-1$ and $t$ will be explained in the following part; for the sake of clarity, the notation of timestep $t$ is omitted unless it is necessary.

\begin{figure*}[th]
\centering
 \includegraphics[clip,width=\linewidth]{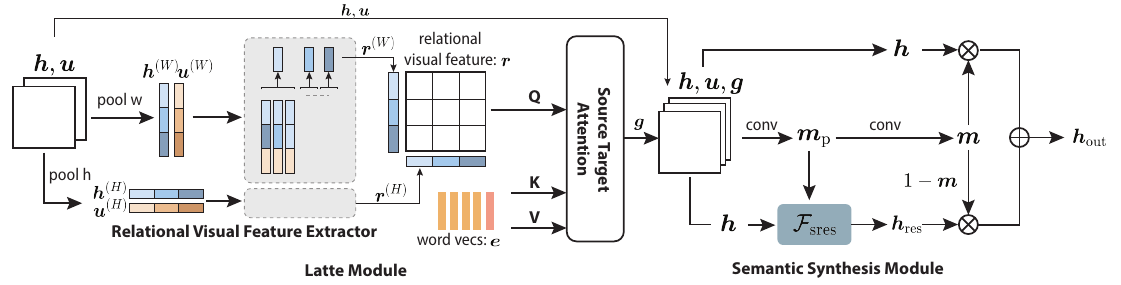}
\caption{\rone{5}{}\textbf{Architecture of the fusion function in the generator}. The fusion function $G_{\text{fusion}}$ consists of the Latte module and semantic synthesis module. The Latte module takes an image feature $\bm{h}$ and a global text feature $\bm{u}$ as inputs, and extracts a fine-grained text-feature vector $\bm{g}$. Following the Latte module is the semantic synthesis module, which takes a concatenation of $\bm{h}$, $\bm{u}$, and $\bm{g}$ as input and yields the modified features $\bm{h}_\mathrm{out}$ as output. }
 \label{fig:model_prop}
\end{figure*}

\subsection{Generator}
\nibf{Overview.} The generator produces a target image $\hat{\bm{x}}_{\mathrm{tgt}} = G(\bm{x}_{\mathrm{src}}, \overline{\bm{e}}, \bm{e}, \bm{z})$, where $\bm{z} \sim \mathcal{N}(0, I) \in \mathbb{R}^{N_z}$ is a noise vector whose size is $N_z$. A sentence embedding $\overline{\bm{e}} \in \mathbb{R}^{E}$, where $E$ is the size of an embedding vector, and a set of word embeddings $\bm{e} = [\bm{e}_i| i = 1, ..., L] \in \mathbb{R}^{L \times E}$, where $L$ is the length of the embeddings, are obtained from an embedding function $f_{\mathrm{text}}$ given a history of instructions as $(\bm{e}, \overline{\bm{e}}) = f_{\mathrm{text}}({x}_{l}^{(t)}, {x}_{l}^{(1:t-1)})$.

More specifically, a computation flow of the generator $G$ is described as follows.
\rone{2}{First, a visual feature map $\bm{h}=G_{\mathrm{image}}(\bm{x}_{\mathrm{src}}) \in \mathbb{R}^{C_h \times H \times W}$} and a text feature map $\bm{u}=G_{\mathrm{text}}(\overline{\bm{e}}, \bm{z}) \in \mathbb{R}^{C_u \times H \times W}$ are computed, where $H$ and $W$ are respectively the height and width of the feature maps, and $C_h$ and $C_u$ are respectively the channel sizes of the feature maps. Then, both visual and textual feature maps are fused to produce a target feature map $\bm{h}_{\mathrm{out}}=G_{\mathrm{fusion}}(\bm{h}, \bm{u}, \bm{e}) \in \mathbb{R}^{C_h \times H \times W}$. Finally, a target image is generated as $\hat{\bm{x}}_{\mathrm{tgt}} = G_{\mathrm{dec}}(\bm{h}_{\mathrm{out}}, \overline{\bm{e}})$. The $G_{\mathrm{fusion}}$, which is the most important block of our model, is made up of the modules described in the rest of this section.

\nibf{Relational Visual Feature Extractor.} A Relational Visual Feature Extractor extracts visual representations, each feature of which includes the relationships with others (Fig.~\ref{fig:model_prop} left). Given a visual feature map $\bm{h}$ and a text feature map $\bm{u}$, width-wise pooled feature vectors $\bm{h}_{i}^{(W)}$ and $\bm{u}_{i}^{(W)}$ at the position $i = \{1, ..., H\}$ over positions $j = \{1, ..., W\}$ are computed as
\begin{align}
 \bm{h}_{i}^{(W)} = \frac{1}{W} \sum_{j=1}^{W} \bm{h}_{ij}, \quad
 \bm{u}_{i}^{(W)} = \frac{1}{W} \sum_{j=1}^{W} \bm{u}_{ij}.
\end{align}
Then, a relational visual representation vector $\bm{r}_{i}^{(W)}$ is obtained by
\begin{align}
 \bm{r}_{i}^{(W)} = \sum_{k=1}^{H} \mathcal{F}_{\mathrm{rel}}
 ([\bm{h}_{i}^{(W)} + \bm{p}_{i}^{(W)}, \bm{h}_{k}^{(W)} + \bm{p}_{k}^{(W)}, \bm{u}_{i}^{(W)}]),
\end{align}
where $\bm{p}_{k}^{(W)}$ is a trainable positional embedding for the $k$-th position and $\mathcal{F}_{\mathrm{rel}}$ consists of linear layers with an activation function. We also obtain  a height-wise pooled relational visual representation vector $\bm{r}_{j}^{(H)}$ in the same way. Finally, we get a relational visual representation map $\bm{r} \in \mathbb{R}^{C_r \times H \times W}$ as follows:
\begin{align}
 \bm{r} &= \bm{r}^{(HW)} \ast W_{r}, \\
 \bm{r}_{ij}^{(HW)} &= [\bm{r}_{j}^{(H)}, \bm{r}_{i}^{(W)}],
\end{align}
where $\ast$ represents the convolution operator with kernel $W_{r}$.

\nibf{Source-target-attention structure.} This structure extracts spatially variant text features that properly reflect the instructions (Fig.~\ref{fig:model_prop} middle). We use a source-target attention structure in which the source is a tuple of word embeddings and a sentence embedding $\tilde{\bm{e}} = (\bm{e}, \overline{\bm{e}})$, and the target is the relational visual representation map $\bm{r}$.

Given $\tilde{\bm{e}} \in \mathbb{R}^{L+1\times E}$, a key-value pair ($K, V$) is generated as follows:
\begin{align}
 K = W_\mathrm{K} \tilde{\bm{e}}^\top, \quad V = W_\mathrm{V} \tilde{\bm{e}}^\top,
 \label{eq:kv}
\end{align}
where $W_{\mathrm{K}} \in \mathbb{R}^{d_{\mathrm{model}} \times E}$ and $W_{\mathrm{V}} \in \mathbb{R}^{d_{\mathrm{model}} \times E}$ are weights for the key and value, respectively, and $d_{\mathrm{model}}$ is a hyper parameter. Given a reshaped relational visual representation map $\tilde{\bm{r}} \in \mathbb{R}^{HW \times C_r}$, a query $Q$ is obtained as
\begin{align}
 Q = W_{\mathrm{Q}} \tilde{\bm{r}}^\top,
\end{align}
where $W_\mathrm{Q} \in \mathbb{R}^{d_{\mathrm{model}} \times C_r}$ is a weight for the query. The output feature is obtained by the scaled-dot-product attention~\citep{vaswani2017attention} as follows:
\begin{align}
 \tilde{\bm{g}} = V \text{softmax}(\frac{K^{\top}Q}{\sqrt{d_{\mathrm{model}}}}),
 \label{eq:sta}
\end{align}
where $d_{\mathrm{model}}$ is a scaling factor. We extend this attention mechanism to a multi-head version (eight heads) as in \citet{vaswani2017attention}. Finally, the output vector $\bm{g} \in \mathbb{R}^{d_{\mathrm{model}} \times H \times W}$ is obtained by reshaping $\tilde{\bm{g}}$.

\nibf{Semantic synthesis module.} Given the image vector $\bm{h}$, the global text feature $\bm{u}$, and the fine-grained text feature vector $\bm{g}$, the semantic synthesis module produces a modified representation $\bm{h}_{\mathrm{out}}$ for image generation (Fig.~\ref{fig:model_prop} right).
Following \citet{vo2019composing} and \citet{kenan2020learning}, we introduce a text-image residual gating structure. 
\rone{4}{The key design idea of this structure is to modify only the area designated in the instruction and leave unrelated areas as they are. }
The modified representation $\bm{h}_\mathrm{out}$ is obtained by applying the gate structure to the residual representation $\bm{h}_{\mathrm{res}}$ and the original representation $\bm{h}$:
\begin{align}
 \bm{h}_{\mathrm{out}} = w_{\mathrm{gate}}(\bm{h} \odot \bm{m})
 + w_{\mathrm{res}}(\bm{h}_{\mathrm{res}} \odot (1 - \bm{m})),
\end{align}
where $\odot$ denotes the Hadamard product, and $w_{\mathrm{gate}}$ and $w_{\mathrm{res}}$ are trainable scaling parameters. In the residual path, the images are normalized in a spatially adaptive manner as follows:
\begin{align}
 \bm{h}_{\mathrm{res}} &= \mathcal{F}_{\mathrm{sres}}(\bm{h}, \bm{m}_{\mathrm{p}}), \\\nonumber
 \bm{m}_{\mathrm{p}} &= [\bm{h}, \bm{u}, \bm{g}] \ast W_1, \\\nonumber
 \bm{m} &= \sigma(\bm{m}_{\mathrm{p}} \ast W_2),
\end{align}
where $\mathcal{F}_{\mathrm{sres}}$ consists of SPADE residual blocks~\citep{park2019semantic}. The gating matrix $\bm{m}$ is generated by applying two convolution operations and the sigmoid activation $\sigma(\cdot)$.

\subsection{Discriminator}
Inspired by \citet{schonfeld2020u}, we introduce the Text-Conditioned U-Net discriminator $D$, which has an image encoder $D_{\mathrm{enc}}$, a local-level unconditional discriminator $D_\mathrm{local}$, and a global-level text-conditional discriminator $D_\mathrm{global}$. Given a real image input ${\bm{x}}_{\mathrm{tgt}}$, the output of the local-level discriminator is
\begin{align}
 \bm{d}_{\mathrm{local}} =
 D_{\mathrm{local}}
 (D_{\mathrm{enc}}({\bm{x}}_{\mathrm{tgt}}))
 ,
 \label{eq:dl}
\end{align}
where $D_{\mathrm{enc}}$ and $D_{\mathrm{local}}$ respectively denote the U-Net encoder and decoder. The shape of $\bm{d}_{\mathrm{local}}$ is the same as the shape of ${\bm{x}}_{\mathrm{tgt}}$ except for the channel size, which is one. The global-level text-conditional discriminator takes a source image $\bm{x}_{\mathrm{src}}$ and the real target image $\bm{x}_{\mathrm{tgt}}$ as inputs and yields a scalar,
\begin{align}
 d_{\mathrm{global}} = D_{\mathrm{global}}(
 (D_{\mathrm{enc}}(\bm{x}_{\mathrm{tgt}}) - D_{\mathrm{enc}}(\bm{x}_{\mathrm{src}})),
 \overline{\bm{e}}
 ),
 \label{eq:dg}
\end{align}
where $D_{\mathrm{global}}$ is the encoder that takes a visual feature difference and a sentence vector $\overline{\bm{e}}$ for projection~\citep{miyato2018cgans}. For discriminating a generated image, $\bm{x}_{\mathrm{tgt}}$ should be replaced by $\hat{\bm{x}}_{\mathrm{tgt}}$ in Eqs.~(\ref{eq:dl}) and (\ref{eq:dg}).

\subsection{Training Objectives}
The training steps are divided into two phases: a pretraining phase and an adversarial training phase. In the pretraining phase, the goal is to tune the text-embedding module $f_{\mathrm{text}}$ by minimizing the $\ell_1$ loss, which is defined as
\begin{align}
 \mathcal{L}_{\mathrm{pretrain}} &= \sum_{(\bm{x}_{\mathrm{src}}, x_l, \bm{x}_{\mathrm{tgt}}) \in \mathcal{D}}
 |f_{\mathrm{image}}(\bm{x}_{\mathrm{tgt}}) - \hat{\bm{h}}_{\mathrm{tgt}}|, \\\nonumber
 \hat{\bm{h}}_{\mathrm{tgt}} &=
 f_{\mathrm{fusion}}(f_{\mathrm{image}}(\bm{x}_{\mathrm{src}}), f_{\mathrm{text}}(x_l)),
\end{align}
where $\mathcal{D}$ is a set of triplets formed by a source image, an instruction, and a target image, and $f_{\mathrm{fusion}}$ consists of a source-target-attention structure and a text-image residual gating structure. Once $f_{\mathrm{text}}$ is pretrained, we freeze its weights  in the following training.

\rtwo{3}{
For the adversarial training, the loss for the discriminator is defined as the sum of two terms:
\begin{align}
 \mathcal{L}_{D}^{\mathrm{(global)}} &= \mathcal{L}_{D_{\mathrm{real}}}
 + \frac{1}{2}(\mathcal{L}_{D_{\mathrm{fake}}} + \mathcal{L}_{D_\mathrm{wrong}})
 + \beta \mathcal{L}_{\mathrm{aux}}, \\\nonumber
 \mathcal{L}_{D}^{\mathrm{(local)}} &= \mathcal{L}_{D_{\mathrm{real}}} + \mathcal{L}_{D_{\mathrm{fake}}},
\end{align}
where $\mathcal{L}_{D_{\mathrm{real}}}$ and $\mathcal{L}_{\mathrm{D}_{\mathrm{fake}}}$ are respective losses for real and generated detection, $\mathcal{L}_{D_{\mathrm{wrong}}}$ is a loss for wrongly matched image-text input pairs of real samples, and $\mathcal{L}_{\mathrm{aux}}$ is an auxiliary loss with a loss-scaling factor $\beta$. We predict the added object in the auxiliary task, which is formalized as multi-label classification. The adversarial loss terms in the discriminator loss are defined as the adversarial hinge loss~\citep{lim2017geometric,miyato2018spectral}. The loss for the generator is defined in the same way as the discriminator loss. More details about the training are provided in the Appendix.}
\begin{table}[t]
\small
\centering
\begin{tabular}{@{}lcc@{}}
\toprule
Statistics                         & CoDraw                 & i-CLEVER             \\ \midrule
Dataset size                       & (7989, 1002, 1002)     & (6003, 2000, 2000)   \\
$N_{\text{turn}}$     & (4.25, 1.46, 14, 1)    & (5, 0, 5, 5)         \\
$N_{\text{token}}$ & (23.05, 11.79, 222, 1) & (14.99, 5.69, 23, 6) \\
Total vocabulary size              & 4090                   & 25                   \\ \bottomrule
\end{tabular}
\vspace{0.5em}
\caption{\rtwo{3}{}
\textbf{Statistics of CoDraw and i-CLEVR datasets.} Dataset size shows the number of images in each training split (train, valid, and test, respectively). $N_{\text{turn}}$ represents the number of turns in a scene in a quadruple representing the average, the standard deviation, the maximum value, and the minimum value, respectively. $N_{\text{token}}$ represents the number of tokens in a sentence in a quadruple as well. Finally, the total vocabulary size shows the size of the vocabulary in each dataset.}
\label{tab:statistics}
\end{table}
\begin{table*}[tb]
\centering
\begin{adjustbox}{width=\textwidth}
\begin{tabular}{@{}lccccccccc@{}}
\toprule
           & \multicolumn{4}{c}{CoDraw}            & \multicolumn{1}{l}{} & \multicolumn{4}{c}{i-CLEVR}           \\ \cmidrule(lr){2-5} \cmidrule(l){7-10} 
Methods      & AP$\uparrow$ & AR$\uparrow$ & F1$\uparrow$ & RSIM$\uparrow$  &
& AP$\uparrow$ & AR$\uparrow$ & F1$\uparrow$ & RSIM$\uparrow$  \\ \midrule

GeNeVA-GAN~{\citep{el2019tell}} & 66.64     & 52.66  & 58.83    & 35.41 &                      & 92.39     & 84.72  & 88.39    & 74.02 \\
GeNeVA-GAN${}^{\dagger}$~{\citep{el2019tell}}  & 54.38     & 54.42  & 54.40    & 38.93 &                      & 71.01     & 42.61  & 53.26    & 30.66 \\ 
SSCR {\citep{fu2020iterative}}
& 58.17 & 56.61 & 57.38 & 39.11 &
& 73.75 & 46.39 & 56.96 & 34.54 \\
TIRG {\cite{kenan2020learning}}
& 76.56     & 73.40  & 72.40    & 46.64  &
& 94.30     & 92.96  & 93.71    & 77.55 \\ \midrule

Ours (w/o Latte \& TxtCond)
& $76.75{\scriptscriptstyle \pm0.94}$    
& $73.99{\scriptscriptstyle \pm0.11}$ 
& $72.81{\scriptscriptstyle \pm0.51}$
& $47.43{\scriptscriptstyle \pm0.28}$  &
& $95.77{\scriptscriptstyle \pm0.59}$
& $94.60{\scriptscriptstyle \pm0.69}$
& $95.08{\scriptscriptstyle \pm0.65}$
& $79.31{\scriptscriptstyle \pm1.23}$ \\

Ours (w/o Latte)
& $79.04{\scriptscriptstyle \pm1.28}$    
& $73.91{\scriptscriptstyle \pm0.53}$ 
& $74.02{\scriptscriptstyle \pm1.03}$
& $48.16{\scriptscriptstyle \pm0.41}$  &
& $98.35{\scriptscriptstyle \pm1.27}$
& $97.68{\scriptscriptstyle \pm1.61}$
& $97.96{\scriptscriptstyle \pm1.47}$
& $82.78{\scriptscriptstyle \pm1.94}$ \\

Ours (w/o TxtCond)
& $80.53{\scriptscriptstyle \pm0.97}$    
& $\bm{78.42}{\scriptscriptstyle \pm0.38}$ 
& $\underline{77.11}{\scriptscriptstyle \pm0.60}$
& $53.13{\scriptscriptstyle \pm0.25}$  &
& $96.67{\scriptscriptstyle \pm0.41}$
& $95.52{\scriptscriptstyle \pm0.43}$
& $95.99{\scriptscriptstyle \pm0.42}$
& $81.04{\scriptscriptstyle \pm0.97}$ \\
\midrule

Ours (Latte w/o RFE) 
& $\underline{80.85}{\scriptscriptstyle \pm0.50}$    
& $78.03{\scriptscriptstyle \pm0.37}$ 
& ${77.05}{\scriptscriptstyle \pm0.40}$
& $\underline{53.87}{\scriptscriptstyle \pm0.30}$  &
& $\underline{98.77}{\scriptscriptstyle \pm0.54}$
& $\underline{98.16}{\scriptscriptstyle \pm0.61}$
& $\underline{98.42}{\scriptscriptstyle \pm0.58}$
& $\bm{84.78}{\scriptscriptstyle \pm1.10}$ \\

Ours (Latte w/o STA) 
& $78.96{\scriptscriptstyle \pm0.46}$    
& $74.37{\scriptscriptstyle \pm0.38}$ 
& $74.20{\scriptscriptstyle \pm0.55}$
& $48.47{\scriptscriptstyle \pm0.24}$  &
& $\bm{99.00}{\scriptscriptstyle \pm0.33}$
& $\bm{98.62}{\scriptscriptstyle \pm0.38}$
& $\bm{98.79}{\scriptscriptstyle \pm0.36}$
& $\underline{84.54}{\scriptscriptstyle \pm0.60}$ \\
\midrule

\bf{Ours (LatteGAN)}
& $\bm{81.50}{\scriptscriptstyle \pm0.82}$    
& $\underline{78.37}{\scriptscriptstyle \pm0.38}$ 
& $\bm{77.51}{\scriptscriptstyle \pm0.52}$
& $\bm{54.16}{\scriptscriptstyle \pm0.21}$  &
& $97.72{\scriptscriptstyle \pm1.44}$
& $96.93{\scriptscriptstyle \pm1.64}$
& $97.26{\scriptscriptstyle \pm1.56}$
& $83.21{\scriptscriptstyle \pm1.70}$ \\ 
\bottomrule

\end{tabular}
\end{adjustbox}
\vspace{0.5em}
\caption{ \textbf{Quantitative comparison and ablation studies.} The methods are compared on the CoDraw and i-CLEVR datasets in accordance with four metrics: the average precision~(AP), average recall (AR), F1-score, and relational similarity (RSIM). For each metric and dataset, the best scores are in bold, and the second best ones are underlined. The mean and standard deviation have been calculated on five different seed runs. }\label{tab:results}
\end{table*}

\sepmode{}
\vspace{-0.5em}
\section{Experiments}
\subsection{Datasets.}
For a fair comparison, we mostly followed the \geneva{} task experiment settings~\citep{el2019tell} and adopted the same two datasets and evaluation metrics\footnote{ All materials, including the datasets and pretrained weights of GeNeVA-GAN, are available online at {\url{https://www.microsoft.com/en-us/research/project/generative-neural-visual-artist-geneva/}}. } .

We used two datasets for the GeNeVA task: CoDraw~\citep{kim2019codraw} and i-CLEVR~\citep{el2019tell}. Both datasets consist of scenes containing a series of images with corresponding instructions. The statistics of the CoDraw and i-CLEVR datasets are presented in Table~\ref{tab:statistics}.

\nibf{CoDraw.} CoDraw is a clip-art-like dataset collected using Amazon Mechanical Turk. The images consist of combinations of 58 unique objects  (such as a boy, girl, and tree) on a static background image consisting of a grassy yard with a blue sky. The object instances in an image can take various poses and sizes that are different from the clip-art template. The instructions consist of conversations between the Teller and Drawer in natural language. In the conversation, the Drawer sequentially updates the image according to the Teller's instructions. The Drawer can ask the Teller a question to clarify an instruction.

\rone{3}{We preprocessed this dataset in the same way as the prior work~\citep{el2019tell}. We concatenated consecutive turns into a single turn until a new object was added or removed by the Drawer. This means that, on every turn, at least one part of an image has been modified from the previous turn.}

\rone{3}{Some instructions are also composed of multiple Teller instructions or Drawer reply messages as a result of the preprocessing. We inserted the special tokens ``{\textless}teller{\textgreater}'' and ``{\textless}drawer{\textgreater}'' at the beginning of each one's sentences. To fix any spelling mistakes , we applied the Bing Spell Check API\footnote{\url{https://azure.microsoft.com/en-us/services/congitive-services/spell-check/}} to all sentences in the dataset. Data splitting was carried out in the same way as in prior works~\citep{kim2019codraw,el2019tell}, with 8K for training, 1K for validation, and 1K for testing.}

\nibf{i-CLEVR.} i-CLEVR is a synthetic dataset generated using the CLEVR~\citep{johnson2017clevr} engine. Each scene contains five image-instruction pairs. Starting from a background image, new objects are added sequentially in a scene. Objects can take several attributes, including three shapes and eight colors. The instructions are generated via a scene graph.
\rtwo{3}{Except on the first turn, which asks to add an object at the center, all instructions in the i-CLEVR dataset include positional referring expressions (e.g., ``in front of the red cube'') and contextual referring expressions such as  ``it''. }
Data splitting was carried out in the same way as in prior work~\citep{el2019tell}, with 6K for training, 2K for validation, and 2K for testing.

\nibf{Implementation details.} The network architecture for the generator and discriminator followed the ResBlocks architecture as used by \citet{miyato2018spectral}. We used instance normalization~\citep{ulyanov2017improved} and AdaIN~\citep{huang2017arbitrary}. For training, we applied teacher forcing by using the ground-truth image $\bm{x}_{v}^{(t-1)}$ instead of the generated image $\hat{\bm{x}}_{v}^{(t-1)}$; however, we used $\hat{\bm{x}}_{v}^{(t-1)}$ during testing. We trained all models until convergence with 400 epochs (or until 72 hours had elapsed) without early-stopping. We chose the weights of the last step to provide the final models. More details are provided in the Appendix.

\subsection{Quantitative Evaluation}
\nibf{Evaluation metrics.} We adopted the same evaluation metrics used in the previous study~\citep{el2019tell} to ensure a fair comparison. The metrics were (i) the object presence matches and (ii) the object position relation matches between the ground-truth and generated images.

The metrics for (i) were the average precision (AP), average recall (AR), and F1 score between the detection results for both the ground-truth and generated images. The AP, AR, and F1 were calculated for each scene. Note that, in the CoDraw and i-CLEVR datasets, more than two instances of the same category do not appear in the same scene. Thus, the object detection could be formulated as a multi-label classification.

As for (ii), we used the relational similarity (RSIM)~\citep{el2019tell}, which evaluates the arrangement of objects. The RSIM is defined as 
\begin{equation}
 \mathrm{RSIM}(E_{G_{\mathrm{gt}}}, E_{G_{\mathrm{gen}}})
 = \mathrm{recall}
 \times \frac{|E_{G_{\mathrm{gt}}} \cap E_{G_{\mathrm{gen}}}|}{|E_{G_{\mathrm{gt}}}|}
 ,
\end{equation}
where ``recall'' indicates the recall over the objects detected in the generated image w.r.t. the objects detected in the ground-truth image. $E_{G_{\mathrm{gt}}}$ is the set of relational edges for the ground-truth image that correspond to vertices that are common to both images. Similarly, $E_{G_{\mathrm{gen}}}$ is the set of relational edges for the generated image that correspond to the vertices in common to both images. We used the last image of each scene to calculate the RSIM.

\begin{figure*}[tbp]
\centering
  \includegraphics[clip,width=\linewidth]{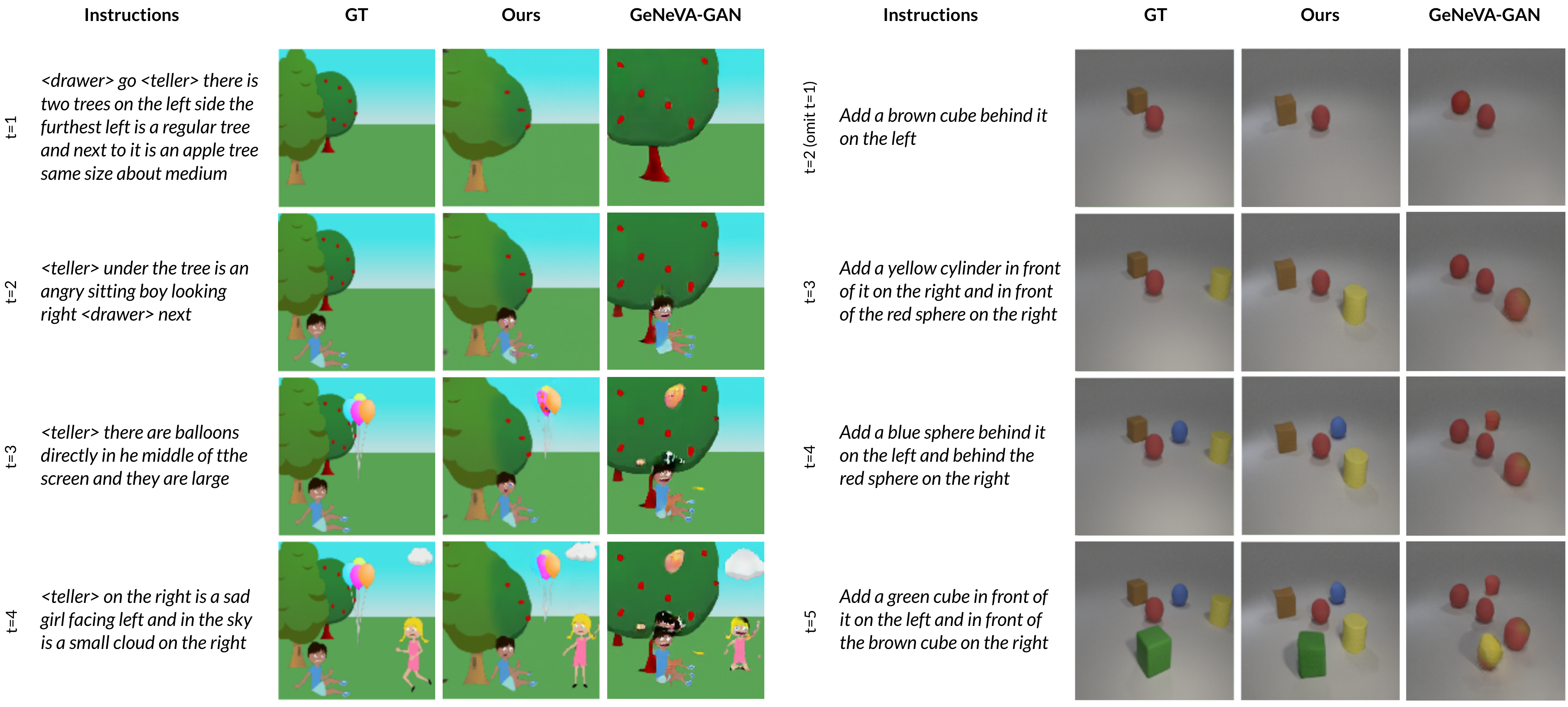}
   \caption{ \textbf{Qualitative comparison for CoDraw~(left) and i-CLEVR~(right) datasets.} Each side shows the instructions, the ground truth samples from the datasets (GT), the generated images by our LatteGAN (Ours), and the generated images by GeNeVA-GAN~\citep{el2019tell}. The generated images were produced in an auto-regressive manner from a background image. We have omitted the images from the first turn for i-CLEVR (``Add a red sphere at the center'') due to space limitations. Note that the generated images can be considered acceptable even if there is a difference from the GT image, due to the ambiguities induced by the instructions. }
  	\label{fig:example}
\end{figure*}

\rtwo{3}{To obtain the detection and localization results for both the ground-truth and generated images, we used a pretrained Inception object detector and localizer available online for CoDraw\footnote{ \url{https://figureqadataset.blob.core.windows.net/live-dataset/GeNeVA-models/codraw_inception_best_checkpoint.pth?st=2019-08-16T20\%3A33\%3A41Z&se=3019-08-17T20\%3A33\%3A00Z&sp=rl&sv=2018-03-28&sr=b&sig=0aheTm73x5pHzwde\%2FRcRBBfrBgVRhE1uxVB4kwk8k7g\%3D}} and for i-CLEVR\footnote{ \url{https://figureqadataset.blob.core.windows.net/live-dataset/GeNeVA-models/iclevr_inception_best_checkpoint.pth?st=2019-08-16T20\%3A34\%3A22Z&se=3019-08-17T20\%3A34\%3A00Z&sp=rl&sv=2018-03-28&sr=b&sig=U9eRRPZHoZDOLO\%2FWYnNAZ9attfFJKlGo28ZX7D\%2BTIDk\%3D}}.}

\nibf{Baseline comparison.} We compared our model with four baselines: GeNeVA-GAN~\citep{el2019tell}, GeNeVA-GAN${}^{\dagger}$, which represents re-implemented results\footnote{ GeNeVA-GAN${}^{\dagger}$ is reported since we also found that the GeNeVA-GAN results could not be reproduced following the official implementation provided on github   ({\url{https://github.com/Maluuba/GeNeVA}}). } reported by~\citep{fu2020iterative}, SSCR~\citep{fu2020iterative}, and TIRG~\citep{kenan2020learning}. Since TIRG is a non-iterative image manipulation model, we made some changes to fit into the iterative scheme; for a fair comparison, we implemented it based on the (w/o Latte\& TxtCond) settings explained in the ablation studies.

Table~\ref{tab:results} quantitatively compares our model with the baselines on the basis of the AP, AR, F1, and RSIM scores on two different datasets. We can see that the F1 and RSIM scores for the CoDraw dataset obtained by LatteGAN were respectively 77.51 and 54.16, which outperformed the baselines at great margins. Our model also outperformed the baselines on the other two metrics. The LatteGAN results for the i-CLEVER dataset were 97.26 for F1 and 83.21 for RSIM scores, which is also the best performance among all compared models.

In summary, our model (LatteGAN) significantly outperformed the baselines on both datasets. In particular, the ARs were greatly improved, which boosted the F1 and RSIM scores (by their definitions). This indicates that our approach has effectively reduced the under-generation of objects.

\nibf{Ablation studies.} As shown in Table~\ref{tab:results}, five ablation conditions were tested. Three of the following ablations were conducted by removing two major components, the Latte module and the Text-Conditioned U-Net Discriminator: (1) a condition without the Latte module (w/o Latte), in which the concatenation of the source image feature $\bm{h}$ and the global text representation $\bm{u}$ was  simply fed to the semantic synthesis function $\mathcal{F}_{\mathrm{SS}}$; (2) a condition in which the Text-Conditioned U-Net discriminator was replaced with a single global text-conditioned discriminator (w/o TxtCond); and (3) a condition combining (1) and (2) (w/o Latte \& TxtCond). Additionally, to provide more insights on the Latte module, we tested two conditions by removing its two sub-components based on (2): (4) a condition without the Relational Feature Module~(RFE) of the Latte Module; and (5) a condition without the Source Target Attention~(STA) of the Latte Module.

Table~\ref{tab:results} shows that the Latte module improved on all of the metrics for CoDraw, especially the AR (by 4.46 points) and the RSIM (by 6.00 points). The improvement of the AR indicates that the Latte module can successfully extract information that cannot be obtained via a global sentence vector. The Text-Conditioned U-Net discriminator improves the precision by 2.46 when it is used along with the Latte module. This indicates that by discriminating images at the local level, the Text-Conditioned U-Net discriminator ensures both the appropriateness of manipulation and the quality of synthesized objects.

As for the i-CLEVR dataset, for which the results were almost saturated, the effect of ablation was less significant. However, the Text-Conditioned U-Net discriminator contributed to the local image quality; as a result, it improved all of the metrics.

The results also show that both the RFE and the STA improved performances. The RFE had a greater improvement on the i-CLEVR, which frequently includes referring expressions, increasing the RSIM by 1.76 (w/o RFE\&STA vs. w/o STA). This indicates the effectiveness of the RFE in extracting relational features. The STA played a more significant role in both datasets, increasing the F1 score by 3.31 (w/o STA vs. LatteGAN) and the RSIM by 5.71 (w/o RFE\&STA vs. w/o RFE) in CoDraw and the RSIM by 2.00 (w/o RFE\&STA vs. w/o RFE) in i-CLEVR. This demonstrates that extracting spatially variant text features is vital in MTIM tasks.

\subsection{Qualitative Evaluation}
Fig.~\ref{fig:example} shows the qualitative results obtained by our model and the baseline on the CoDraw and i-CLEVR datasets, displayed in tiles. The left, middle, and right columns show the ground-truth images, the images generated by our model, and those generated by the baseline, respectively.

In the fourth row ($t=4$) of the CoDraw results, the Drawer was instructed to draw ``a sad girl facing left''; however, the baseline generated an image of a girl facing right, while our model successfully generated a correct image. Moreover, the baseline did not clearly draw the boy's head and legs at $t=3$, and the image collapsed. In comparison, our model generated better images in general, except for $t=1$, where the apple tree trunk was not drawn.

In the first row ($t=2$) of the i-CLEVR results, the Drawer was instructed to draw ``a brown cube''; however, the baseline generated a red sphere, while our model successfully generated an image of a brown cube at the desired position. Similarly, the baseline added wrong objects (red spheres) for $t=3$ and $t=4$, while our model successfully added the instructed objects at every step.

\begin{figure}[t]
  \begin{center}
    \includegraphics[clip,width=\linewidth]{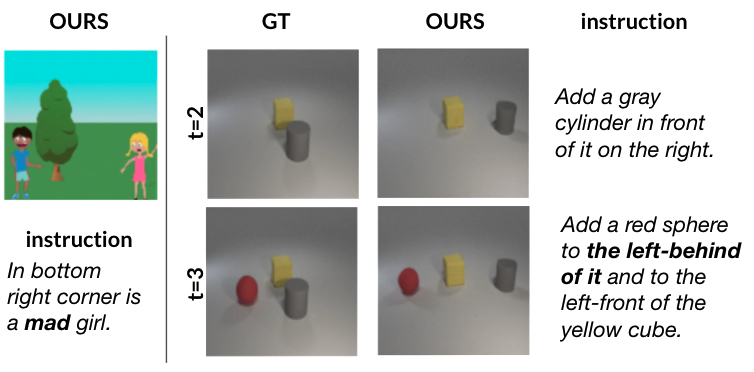}
  \vspace{-1em}
   \caption{\rtwo{3}{}\textbf{Example  failures of LatteGAN.} CoDraw: Failed to generate a facial expression as instructed; a surprised face was generated instead. i-CLEVR: At $t=3$, the red sphere was placed in a slightly wrong position because the instruction at $t=2$ was ambiguous. }\label{fig:failed}
  \end{center}
\end{figure}
\rtwo{3}{Fig.~\ref{fig:failed} presents examples of failed samples produced by LatteGAN. In CoDraw, LatteGAN tended to fail to generate details, such as the posture and facial expressions of people. In i-CLEVR, LatteGAN failed in a very limited situation where it followed the past instructions correctly but could not follow the next because there was no space left to add an object. This is an inevitable error that stems from the data including ambiguous instructions. However, we believe we can mitigate this problem in the future by retaining multiple generated results and selecting an appropriate image on each iteration.}

\sepmode{}

\section{Conclusion}
We have presented \prop{}, a novel GAN architecture for MTIM. The Latte module is implemented to feed the fine-grained text representations to the generator. We also proposed a Text-Conditional U-Net discriminator for handling not only text-conditioned global representations but also text-unconditioned local representations to improve the distinguishability of generated objects. Experimental results demonstrated that our model outperformed several baselines  on both the CoDraw and i-CLEVR datasets.

\newpage
\appendix
\section*{APPENDIX}
\renewcommand{\thesubsection}{\Alph{subsection}}
\subsection{Training Details}
\nibf{Loss of the discriminator.} The loss for the discriminator is defined as follows:
\begin{align}
 \mathcal{L}_{D} &= \mathcal{L}_{D}^{\mathrm{(global)}} + \mathcal{L}_{D}^{\mathrm{(local)}}, \label{eq:d}\\\nonumber
 \mathcal{L}_{D}^{\mathrm{(global)}} &= \mathcal{L}_{D_{\mathrm{real}}}
 + \frac{1}{2}(\mathcal{L}_{D_{\mathrm{fake}}} + \mathcal{L}_{D_\mathrm{wrong}})
 + \beta \mathcal{L}_{\mathrm{aux}}, \\\nonumber
 \mathcal{L}_{D}^{\mathrm{(local)}} &= \mathcal{L}_{D_{\mathrm{real}}} + \mathcal{L}_{D_{\mathrm{fake}}},
\end{align}
where the auxiliary loss function for $\mathcal{L}_{\mathrm{aux}}$ is defined as a binary cross entropy over the added object at the current timestep. The adversarial loss terms in Eq.~(\ref{eq:d}) are defined as the adversarial hinge loss~\citep{lim2017geometric,miyato2018spectral} as follows:
\begin{align}
 \mathcal{L}_{D_{\mathrm{real}}} &= -\mathbb{E}_{(\bm{x}_v, x_l) \sim p_{\mathrm{data}}}
 \left[\mathrm{min}(0, -1 + D(\bm{x}_v, x_l))\right], \\
 \mathcal{L}_{D_{\mathrm{fake}}} &= -\mathbb{E}_{z \sim \mathcal{N}, x_l \sim p_{\mathrm{data}}}
 \left[\mathrm{min}(0, -1 - D(G(z, x_l), x_l))\right], \\
 \mathcal{L}_{D_{\mathrm{wrong}}} &= -\mathbb{E}_{(\bm{x}_v, \hat{x}_l) \sim p_{\mathrm{data}}}
 \left[\mathrm{min}(0, -1 - D(\bm{x}_v, \hat{x}_l))\right],
\end{align}
where $D$ and $G$ are respectively a discriminator and generator function, $p_\mathrm{data}$ is a dataset, $\mathcal{N}$ is the standard Gaussian distribution, and $\hat{x}_l$ is a wrong instruction. In computing $\mathcal{L}_D^{\mathrm{(global)}}$, a discriminator $D$ will return its decision at the global level, while in $\mathcal{L}_D^{\mathrm{(local)}}$, it will return its decision at the local level~\cite{schonfeld2020u}.

\nibf{Loss of the generator.} The loss for the generator is defined as
\begin{align}
 \mathcal{L}_{G} &= \mathcal{L}_{G}^{\mathrm{(global)}} + \mathcal{L}_{G}^{\mathrm{(local)}}, \label{eq:g}\\\nonumber
 \mathcal{L}_{G}^{\mathrm{(global)}} &= \mathcal{L}_{G_{\mathrm{fake}}} + \beta \mathcal{L}_{\mathrm{aux}}, \\\nonumber
 \mathcal{L}_{G}^{\mathrm{(local)}} &= \mathcal{L}_{G_{\mathrm{fake}}},
\end{align}
where $\mathcal{L}_{G_{\mathrm{fake}}}$ is the adversarial hinge loss
\begin{align}
 \mathcal{L}_{G_{\mathrm{fake}}} = -\mathbb{E}\left[d_{\mathrm{fake}}\right],
\end{align}
where the scalar $d_{\mathrm{fake}}$ corresponds to the output of the discriminator given a generated sample, and belongs to $\bm{d}_{\mathrm{local}}$ or $d_{\mathrm{global}}$ depending which loss ($\mathcal{L}_{G}^{\mathrm{(global)}}$ or $\mathcal{L}_{G}^{\mathrm{(local)}}$) is computed.

\nibf{Regularization terms.} To increase the stability of the adversarial training, we apply two regularization terms to the additional loss terms in Eqs.~(\ref{eq:d}, \ref{eq:g}). One is the zero-centered gradient penalty~\citep{mescheder2018training}, which is added to the discriminator loss of Eq.~(\ref{eq:d}). It is applied to the discriminator $D$ parameters  $\psi$, considering only the real target data:
\begin{align}
 \mathcal{L}_{\mathrm{reg}} = \frac{\gamma}{2} \mathbb{E}
 \left[\| \nabla D_{\psi}({\bm{x}}_{\mathrm{tgt}}) \|^2\right],
\end{align}
where $\gamma$ is a weighting factor. The other is for a conditioning augmentor~\citep{zhang2017stackgan}. Its loss is added to the generator loss of Eq.~(\ref{eq:g}). The conditioning augmentor is a module that samples text-conditioned latent variable $\bm{c}$ from independent Gaussian distribution $\mathcal{N}(\mu(\overline{\bm{e}}), \Sigma(\overline{\bm{e}}))$, where the mean $\mu(\overline{\bm{e}})$ and diagonal covariance matrix $\Sigma(\overline{\bm{e}})$ are functions of the sentence embedding $\overline{\bm{e}}$. The text-conditioned latent variable $\bm{c}$ is used as the input of $G_{\mathrm{text}}$ instead of the sentence embedding $\overline{\bm{e}}$. The Kullback-Leibler (KL) divergence is introduced as a regularization term in order to increase smoothing on the conditioning manifold, as
\begin{align}
 \mathcal{L}_{\mathrm{ca}} = \alpha D_{KL}(\mathcal{N}(\mu(\overline{\bm{e}}), \Sigma(\overline{\bm{e}})) \| \mathcal{N}(0, I)),
\end{align}
where $\mathcal{N}(0, I)$ is the standard Gaussian distribution and $\alpha$ is a weighting factor.

\footnotetext[3]{\url{https://huggingface.co/transformers/model_doc/bert.html##transformers.BertTokenizerFast}}

\subsection{Implementation Details}
\nibf{Transforms.} All images were resized to $(128, 128)$ and normalized between $(-1, 1)$. Instructions were tokenized using the uncased BERT tokenizer available online\footnotemark[3]. When training the GAN model, all instructions were transformed in advance into text embeddings and word embeddings by using the fine-tuned BERT sentence encoder. A small amount of noise sampled from a uniform distribution $\mathrm{Uniform}(0, 1/64)$ was added to the image at every step to stabilize the GAN training dynamics.

\nibf{Training.} We used spectral normalization~\citep{miyato2018spectral} for all layers in both the generator and discriminator. We used the Adam~\citep{kingma2014adam} optimizer with learning rate of 0.0004 for the discriminator and 0.0001 for the generator and betas of (0.0, 0.9) for both. We prepared the evaluation-specific generator and updated its weights by using an exponential moving average of the generator weights during training~\citep{brock2018large}. We used the truncation trick for the latent noise with a threshold of 2.0 during testing~\citep{brock2018large}.

\nibf{Hardware.} The training process took 72 hours with four  NVIDIA Tesla V100 (SXM2) GPUs, two  Intel Xeon Gold 6148 (27.5M cache, 2.40 GHz, 20 cores) processors, and 384 GiB of memory. The average inference time per batch of ten samples was 0.76 s for CoDraw and 0.91 s for i-CLEVR.

\subsection{Supplemental Results}
\begin{figure}[t]
 \begin{center}
 \includegraphics[clip,width=0.85\linewidth]{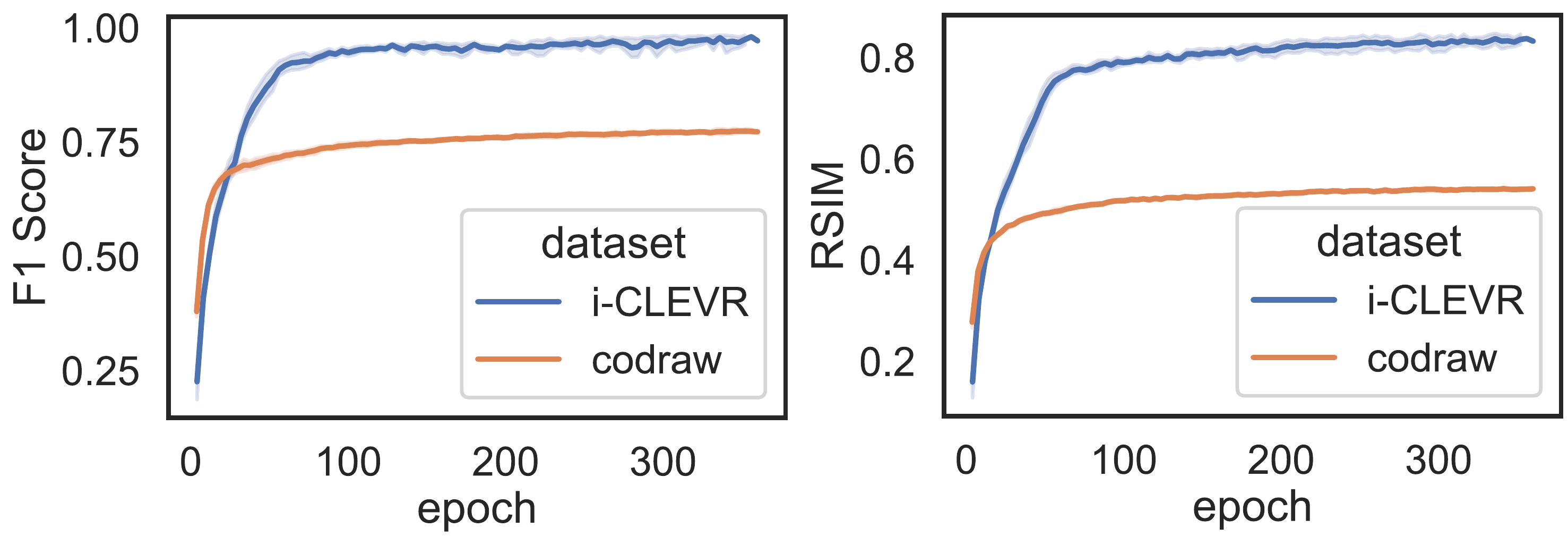}
\caption{\textbf{Learning curve of LatteGAN.}}\label{fig:curve}
 \end{center}
\end{figure}

The convergence curve of LatteGAN for the F1 and the RSIM is presented in Fig.~\ref{fig:curve}.

\begin{figure*}[tbp]
\centering
 \includegraphics[clip,width=\linewidth,height=0.95\textheight]{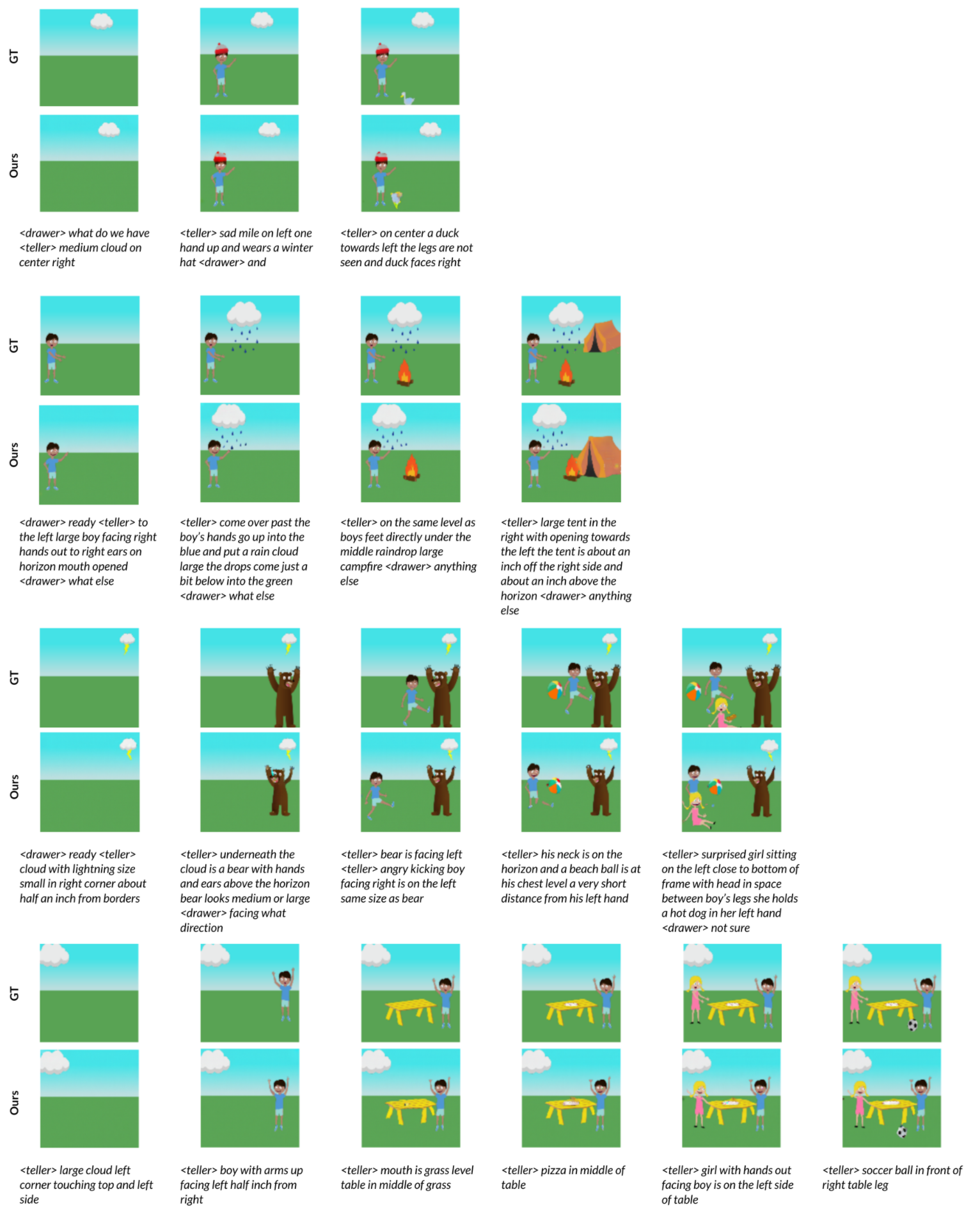}
\caption{Additional generated images from our LatteGAN for the CoDraw dataset.}
\label{fig:adda1}
\end{figure*}

\begin{figure*}[tbp]
\centering
 \includegraphics[clip,width=\linewidth,height=0.95\textheight]{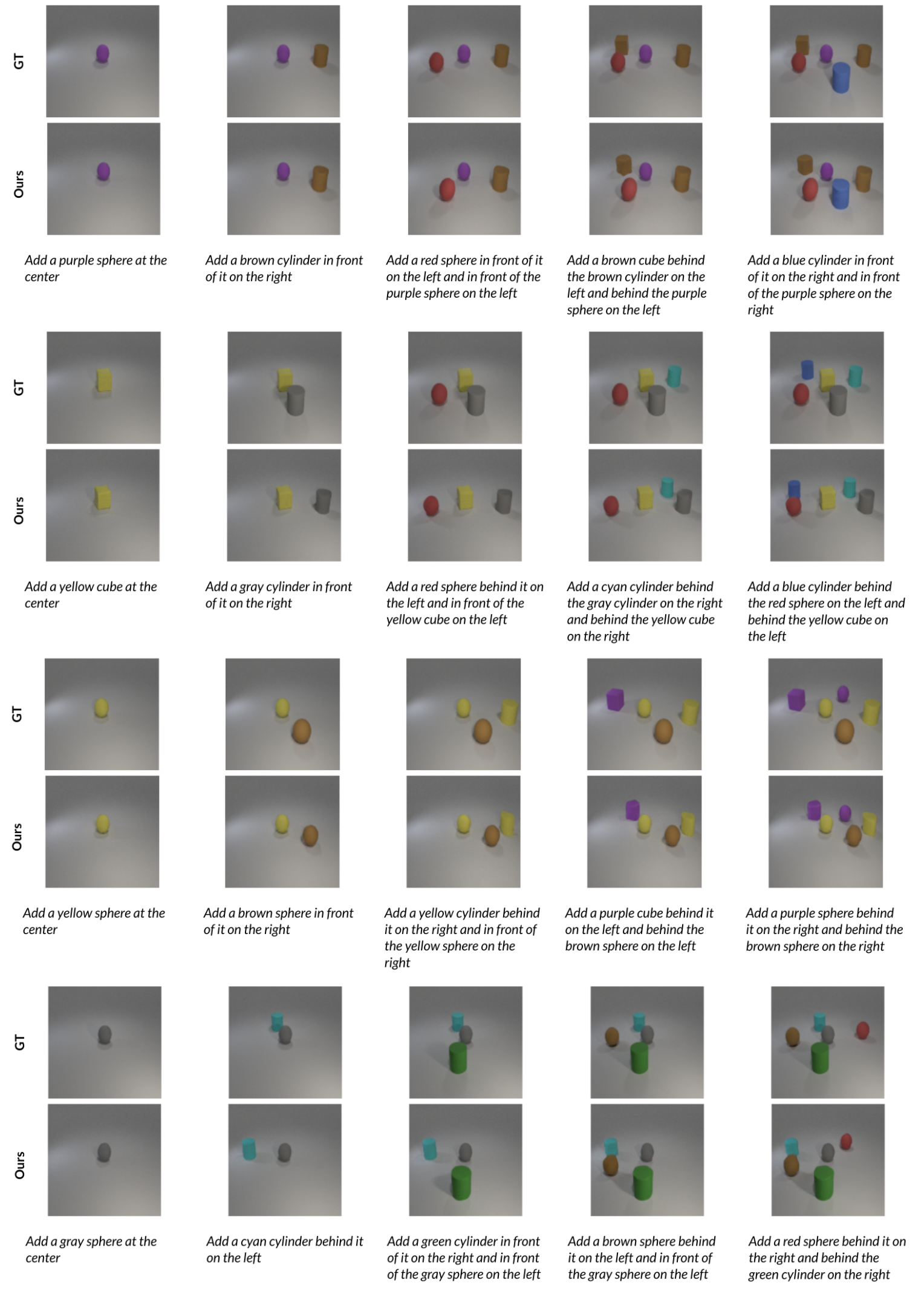}
\caption{Additional generated images from our LatteGAN for the i-CLEVR dataset.}
\label{fig:adda2}
\end{figure*}

Additional images generated from our LatteGAN are presented in Fig.~\ref{fig:adda1} for the CoDraw dataset and in Fig.~\ref{fig:adda2} for the i-CLEVR dataset.

\newpage
\section*{Acknowledgements}
This work was supported by JST CREST Grant Number JPMJCR19A1, Japan, and JSPS KAKENHI Grant Number JP21J13789, Japan.
The experiments were partially conducted on the AI Bridging Cloud Infrastructure (ABCI) provided by the National Institute of Advanced Industrial Science and Technology (AIST). The training progress was tracked and the reports were created with Weight \& Biases. Finally, we thank our colleagues, Nathan Boyer and David Felices, for helpful comments and discussion.


\clearpage
{
\bibliographystyle{IEEEtranN.bst}
\bibliography{biblio}
}

\end{document}